\pdfoutput=1

\documentclass[11pt]{article}

\usepackage[final]{acl}

\usepackage{times}
\usepackage{latexsym}
\usepackage{hyperref}
\usepackage{url}
\usepackage{amsmath}
\usepackage{amssymb}
\usepackage{mathrsfs}
\usepackage{graphicx}
\usepackage{subcaption}
\usepackage{booktabs}
\usepackage{multirow}
\usepackage{algorithm}
\usepackage{algpseudocode}
\usepackage{todonotes}
\usepackage{pifont}

\usepackage[T1]{fontenc}

\usepackage[utf8]{inputenc}

\usepackage{microtype}

\usepackage{inconsolata}

\title{Full Parameter Fine-tuning for Large Language Models \\with Limited Resources}

\author{Kai Lv\textsuperscript{1,2},
Yuqing Yang\textsuperscript{1}, 
Tengxiao Liu\textsuperscript{1}, Qinghui Gao\textsuperscript{1}, 
Qipeng Guo\textsuperscript{2}\thanks{Corresponding author.},
Xipeng Qiu\textsuperscript{1} \\
\textsuperscript{1}School of Computer Science, Fudan University\\
\textsuperscript{2}Shanghai AI Laboratory\\
\texttt{\{klv21, yuqingyang21, txliu21\}@m.fudan.edu.cn} \\
\texttt{guoqipeng@pjlab.org.cn},
\texttt{xpqiu@fudan.edu.cn}
}
\begin{document}
\maketitle
\begin{abstract}
Large Language Models (LLMs) have revolutionized Natural Language Processing (NLP) but demand massive GPU resources for training. Lowering the threshold for LLMs training would encourage greater participation from researchers, benefiting both academia and society.
While existing approaches have focused on parameter-efficient fine-tuning, which tunes or adds a small number of parameters, few have addressed the challenge of tuning the full parameters of LLMs with limited resources. In this work, we propose a new optimizer, \textbf{LO}w-\textbf{M}emory \textbf{O}ptimization (\textbf{LOMO}), which fuses the gradient computation and the parameter update in one step to reduce memory usage. By integrating LOMO with existing memory saving techniques, we reduce memory usage to 10.8\% compared to the standard approach (DeepSpeed solution). Consequently, our approach enables the full parameter fine-tuning of a 65B model on a single machine with 8$\times$RTX 3090, each with 24GB memory.\footnote{Code and data are available at \url{https://github.com/OpenLMLab/LOMO}.}
\end{abstract}

\section{Introduction}
Large Language Models (LLMs) have revolutionized Natural Language Processing (NLP), demonstrating remarkable abilities such as emergence and grokking~\citep{DBLP:journals/tmlr/WeiTBRZBYBZMCHVLDF22}, pushing model size to become larger and larger. However, training these models with billions of parameters, such as those with 30B to 175B parameters, raises the bar for NLP research. Tuning LLMs often requires expensive GPU resources, such as 8$\times$80GB devices, making it difficult for small labs and companies to participate in this area of research.

Recently, parameter-efficient fine-tuning methods~\citep{DBLP:journals/corr/abs-2203-06904}, such as LoRA~\citep{DBLP:conf/iclr/HuSWALWWC22} and Prefix-tuning~\citep{DBLP:conf/acl/LiL20}, provide solutions for tuning LLMs with limited resources. However, these methods do not offer a practical solution for full parameter fine-tuning, which has been acknowledged as a more powerful approach than parameter-efficient fine-tuning~\citep{DBLP:journals/corr/abs-2203-06904, DBLP:journals/corr/abs-2304-08109}. In this work, we aim to explore techniques for accomplishing full parameter fine-tuning in resource-limited scenarios.

\begin{figure*}[t]
    \centering
    \includegraphics[width=0.95\linewidth]{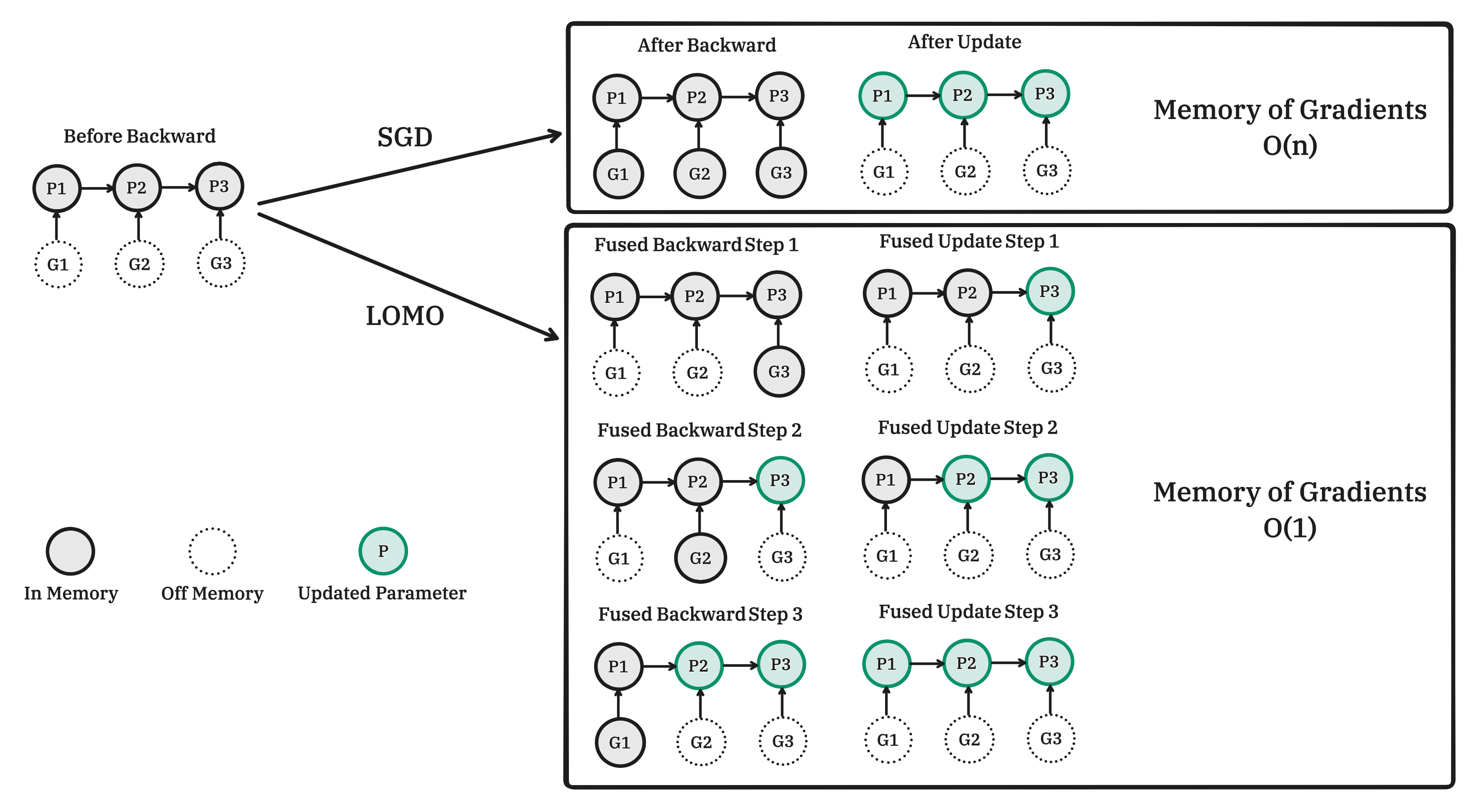}
    \caption{Comparison of SGD and LOMO in backpropagation and parameter update stages. $\textbf{Pi}$ refers to the parameter of the model and $\textbf{Gi}$ refers to the gradient corresponding to $\textbf{Pi}$. LOMO fused gradient computation and parameter update in one step to minimize the size of gradient tensors.}
    \label{fig:inplace_sgd}
\end{figure*}

We analyze the four aspects of memory usage in LLMs, namely activation, optimizer states, gradient tensor and parameters, and optimize the training process in three folds: 1) We rethink the functionality of an optimizer from an algorithmic perspective and find that SGD is a good replacement in terms of fine-tuning full parameter for LLMs. This allows us to remove the entire part of optimizer states since SGD does not store any intermediate state (Sec-\ref{sec:sgd}). 2) Our proposed optimizer, LOMO as illustrated in Figure~\ref{fig:inplace_sgd}, reduces the memory usage of gradient tensors to $O(1)$, equivalent to the largest gradient tensor's memory usage (Sec-\ref{sec:inplace_update}). 3) To stabilize mix-precision training with LOMO, we integrate gradient normalization, loss scaling, and transition certain computations to full precision during training (Sec-\ref{sec:stablize_training}).

Our technique results in memory usage that equals the usage of parameters plus activation and the largest gradient tensor.
We push the memory usage of full parameter fine-tuning to an extreme, making it merely equivalent to the usage of inference. This is because the memory usage of the forward + backward process should not be less than the forward process alone. It is worth noting that, when employing LOMO to save memory, we ensure that the fine-tuning process remains uncompromised, as the parameter update process is still equivalent to SGD.

We empirically assess the memory and throughput performance of LOMO and show that the usage of LOMO enables successful training of a 65B model with only 8 RTX 3090 GPUs. Additionally, to validate the downstream performance of our proposed technique, we apply LOMO to tune the full parameters of LLMs on the SuperGLUE dataset collection~\citep{DBLP:conf/nips/WangPNSMHLB19}. The empirical results demonstrate the efficiency and effectiveness of LOMO for optimizing LLMs with billions of parameters. Overall, our contributions are as follows:

\begin{itemize}
    \item We provide a theoretical analysis suggesting that SGD can successfully fine-tune the full parameters of LLMs. The issues that previously hindered the widespread usage of SGD may no longer be severe problems for fine-tuning LLMs.
    \item We propose LOw-Memory Optimization, named LOMO, to significantly save GPU memory usage without harming the fine-tuning process.
    \item Through a thorough evaluation of memory usage and throughput performance, we empirically validate the effectiveness of LOMO in optimizing LLMs under resource-constrained scenarios. This is further supported by performance evaluations on downstream tasks.
\end{itemize}

\section{Related Work}
In this section, we present related work on memory-saving techniques during full parameter fine-tuning. These techniques can be effectively combined with LOMO to further reduce memory consumption.

\paragraph{Activation Checkpointing} During vanilla backpropagation, all activations from the forward pass are stored in memory to compute gradients. 
This can be a significant memory overhead, especially for large language models. Alternatively, one could discard all activations and recompute them on demand for gradients computation in order to save memory. However, this can result in a substantial additional computation cost. Activation checkpointing  (or gradient checkpointing) takes into account both memory usage and computational cost, providing a compromise solution~\citep{chen2016training}. The activations of strategically selected checkpoint nodes in the computational graph are kept in memory after the forward pass, while the activations of  remaining nodes are recomputed at most once. The activation memory can be reduced to the square root of the original amount at the cost of one extra forward pass.

\paragraph{Mixed-Precision Training} Mixed-precision training has become a prevalent approach for training large language models due to its ability to accelerate training speed and reduce memory footprint~\citep{Megatron-LM,Zero}. By employing half-precision storage for parameters, activations, and gradients, mixed-precision training enables high-throughput computations.
In order to uphold stability and model accuracy, \citet{micikevicius2018mixed}  proposed three techniques which involve the use of full precision copies of weights, loss scaling, and the execution of specific arithmetic operations in full precision.

\paragraph{Heterogeneous Training System} Multiple studies~\citep{7783721,article,9407112} have attempted to reduce GPU memory consumption by leveraging heterogeneous memory, such as CPU and NVMe memory. L2L~\citep{pudipeddi2020training} employs a layer-to-layer strategy, where only the tensors necessary for the computation of a particular layer are transferred to the GPU memory, while the remaining tensors are retained in the CPU memory.
ZeRO-Offload~\citep{Zero-offload}, an extension of ZeRO-2~\citep{Zero}, reserves the gradients and optimizer states in the CPU memory and updates parameters through CPU computation. 
ZeRO-Infinity~\citep{Zero-infinity}, a subsequent advancement of ZeRO-Offload on ZeRO-3~\citep{Zero}, enables further scaling of the model size. 

In addition to the methods orthogonal to LOMO mentioned above, recent developments have introduced several memory-efficient optimization techniques. MeZO~\citep{DBLP:journals/corr/abs-2305-17333} employs a zero-order optimization approach, estimating gradients using two forward passes and updating parameters in place. GaLore~\citep{DBLP:journals/corr/abs-2403-03507} performs low-rank decomposition on gradients and uses these approximated gradients for parameter updates. Other approaches reduce memory usage by quantizing optimizer states~\citep{bit8optim,bit4optim}. Compared to these methods, LOMO neither approximates gradients nor requires low-bit quantization.

\section{Method}

\subsection{Rethink the Functionality of Optimizer} \label{sec:sgd}
The optimizer states occupy a large part of the memory used for training LLMs. Modern optimizer like Adam~\citep{DBLP:journals/corr/KingmaB14} stores intermediate states that are twice the size of parameters. As the size of parameters increases, the optimizer states become the dominant term of memory usage.

\subsubsection{Using SGD}

Although Adam has achieved great success in training deep models, we ask the question ``Can we use a cheaper optimizer for fine-tuning LLMs?" Our answer is SGD, the most basic optimizer. Fortunately, we find that it is an acceptable solution for fine-tuning LLMs when we limit the scope.

Prior works often discuss three challenges of SGD: 1) large curvature loss surface, 2) local optimum, and 3) saddle points~\citep{DBLP:journals/corr/Ruder16,DBLP:journals/tcyb/SunCZZ20}. Modern optimizers have shown effectiveness in dealing with the 1) problem and can mitigate 2) and 3) in some cases. However, when we limit the scope to fine-tuning LLMs, these three challenges could be different.

\paragraph{Smoother loss surface} One important assumption is that the parameter space of LLMs is quite smooth and small perturbations on the parameters will not change the loss too much. There are empirical results and theoretical analyses supporting this assumption~\citep{DBLP:conf/emnlp/HaoDWX19}. If we believe that larger models have a smoother loss surface, we can conclude that the 1) problem is not an issue since the loss surface of LLMs should not have a large curvature. Note that this holds only when we teach the LLMs natural language-based tasks (or code-based if pre-trained with code). A synthetic loss function unrelated to pre-training tasks will indeed face the large curvature problem.

\paragraph{Local optimum is good enough} The goal of fine-tuning is to adapt LLMs to new tasks and domains without significantly changing the model itself. Therefore, a local optimum is often a good enough solution~\cite{DBLP:journals/neco/KawaguchiHK19a}, and the limited training data (compared to pre-training corpus) makes it difficult to push the model to a faraway global optimum.

\paragraph{Distant saddle points} Similarly, for a common NLP task, the initial point of LLMs should be in a valley. If the model is pre-trained with instructions (tasks), the phenomenon could be much more apparent since we have more chances of finding pre-trained tasks that are similar to the new task. Saddle points typically appear on ridges and have a distance from valleys, so we may not encounter the saddle point problem if we do not change the parameter too far from the pre-trained value.

However, there is no guarantee that SGD is a powerful optimizer compared to modern optimizers. Our intuition is to create a simple and practical solution for fine-tuning LLMs and identify its flaws to continually improve it.

\subsubsection{Implicit Batch Size}
Besides the above qualitative discussion, we want to provide a deeper analysis of the stability of fine-tuning LLMs with SGD. Suppose we have a pre-trained model $f(\cdot)$ with the parameter $\boldsymbol{\theta}$, a training set $\mathcal{D}=\{d_1, d_2, \cdots, d_n\}$, and a loss function $\mathcal{L}$. One step update of SGD on a batch with two data points could be,
\begin{gather}
    \boldsymbol{\theta'} = \boldsymbol{\theta} - \alpha  [\nabla \mathcal{L}(d_i, f(d_i, \boldsymbol{\theta})) + \nabla \mathcal{L}(d_j, f(d_j, \boldsymbol{\theta}))], \label{eq:bs2}
\end{gather}
where $\alpha$ is the learning rate, and $d_i, d_j$ are two different training samples. 

Next, two steps update of SGD on these two training samples $d_i, d_j$ sequentially could be, 
\begin{align}
    \boldsymbol{\theta_1} =&\ \boldsymbol{\theta} - \alpha  \nabla \mathcal{L}(d_i, f(d_i, \boldsymbol{\theta})), \\
    \boldsymbol{\theta_2} =&\ \boldsymbol{\theta_1} - \alpha  \nabla \mathcal{L}(d_j, f(d_j, \boldsymbol{\theta_1})).
\end{align}

By differential mean value theorem, we have
\begin{align}
    &\mathcal{L}(d_j, f(d_j, \boldsymbol{\theta_1})) = \mathcal{L}(d_j, f(d_j, \boldsymbol{\theta})) \notag \\
    &+ \nabla\mathcal{L}(d_j, \xi)(f(d_j, \boldsymbol{\theta_1})
    -f(d_j, \boldsymbol{\theta})), \\
    &\boldsymbol{\theta_2} = \boldsymbol{\theta} - \alpha  \nabla \mathcal{L}(d_i, f(d_i, \boldsymbol{\theta})) \notag \\
    &- \alpha \nabla  \mathcal{L}(d_j, f(d_j, \boldsymbol{\theta})) \notag \\
    &- \alpha \nabla [\nabla\mathcal{L}(d_j, \xi)(f(d_j, \boldsymbol{\theta_1})-f(d_j, \boldsymbol{\theta}))], \\ 
    &\boldsymbol{\theta_2} = \boldsymbol{\theta} - \alpha [ \nabla \mathcal{L}(d_i, f(d_i, \boldsymbol{\theta})) \notag \\
    &+ \nabla  \mathcal{L}(d_j, f(d_j, \boldsymbol{\theta}))] \notag \\
    &- \alpha \nabla [\nabla\mathcal{L}(d_j, \xi)(f(d_j, \boldsymbol{\theta_1})-f(d_j, \boldsymbol{\theta}))], \label{eq:bs1}
\end{align}      
where $\xi$ is a point between $f(d_j, \boldsymbol{\theta})$ and $f(d_j, \boldsymbol{\theta_1})$, and we can see that \eqref{eq:bs1} minus \eqref{eq:bs2} equals the $\alpha \nabla [\nabla\mathcal{L}(d_j, \xi)(f(d_j, \boldsymbol{\theta_1})-f(d_j, \boldsymbol{\theta}))]$. Suppose the loss surface is smooth enough, this term become negligible. It suggests that utilizing SGD optimizer over a smooth loss surface could imply a larger batch size. 

As we mentioned above, it's reasonable to assume that the loss surface of LLMs is smooth, and a larger batch size indicates stronger training stability, so we believe that finetuning process of LLMs with the SGD optimizer is stable. This also explains why SGD failed on small models but worked for large models. 
\subsection{LOMO: LOw-Memory Optimization} \label{sec:inplace_update}

\begin{algorithm}[t]
\caption{Fusion Update in LOMO}\label{alg:inplace_sgd}
\begin{algorithmic}[1]
\Require model $f(\cdot)$ with $L$ layers and $p$ parameters, parameter $\boldsymbol{\theta} \in \mathbb{R}^{p}$ , learning rate $\alpha$, max step $T$, training dataset $\mathcal{D}$, loss function $\mathcal{L}$
\vspace{0.1cm}
\For{$t = 1, \dots, T$}
    \State Sample batch $\mathcal{B}=(\boldsymbol{x},\boldsymbol{y}) \subset \mathcal{D}$
    \State $\boldsymbol{\hat{y}} \gets f(\boldsymbol{x},\boldsymbol{\theta})$ \Comment{Forward pass}
    \State $\ell \gets \mathcal{L}(\boldsymbol{y},\boldsymbol{\hat{y}})$
    
    \For{$l = L,\dots,1$} \Comment{Backward}
        \State $\boldsymbol{\theta_{l}} \gets \left[ \theta_i\  \mathrm{for}\ \theta_i \in \mathrm{layer} \ l \right]$ 
        \vspace{0.03cm}
        \State $\boldsymbol{g_l} \gets \frac{\partial \ell}{\partial \boldsymbol{\theta_{l}}}$
        \vspace{0.03cm}
        \State $\boldsymbol{\theta_{l}} \gets \boldsymbol{\theta_{l}} - \alpha * \boldsymbol{g_l}$
        \State $\boldsymbol{g_l} \gets$ None \Comment{Clear gradients}
    \EndFor
    
\EndFor
\end{algorithmic}
\end{algorithm}

The gradient tensor represents the gradient of a parameter tensor and has the same size as the parameter, resulting in a large memory overhead. Modern deep learning training frameworks like PyTorch~\citep{paszke2017automatic} store gradient tensors for all parameters. There are two reasons for storing gradient tensors: computing optimizer states and normalizing gradients.

Since we take SGD as the optimizer, there are no optimizer states relying on gradients, and we have some alternatives to gradient normalization. Thus, we proposed LOw-Memory Optimization (LOMO) as illustrated in Algorithm \ref{alg:inplace_sgd}, fusing the gradient computation and parameter update in one step to avoid storing any gradient tensors. 

In detail, we can express the vanilla gradient descent as $\mathrm{grad} = \frac{\partial \mathcal{L}}{\partial p}, p = p - lr * \mathrm{grad}$, which is a two-step process, computing the gradients first and updating it to the parameters. The fusion version is $p = p - lr * \frac{\partial \mathcal{L}}{\partial p}$.

The key idea is to update the parameter immediately when its gradient is computed so that we do not store gradient tensor in memory. This can be achieved by injecting hook functions into the backward propagation.\footnote{We should inject different hook functions accordingly if some of them share the weight. } PyTorch provides relevant APIs for injecting hook functions, but we cannot implement the exact immediate update with current APIs. Instead, we store at most one parameter's gradient in memory and update each parameter one by one along with the backward propagation. Our approach reduces the memory usage of gradients from storing of all parameters' gradients to storing only one parameter's gradient. 

The majority of LOMO memory usage coincides with that of parameter-efficient fine-tuning (PEFT) methods, indicating that combining LOMO with these methods only introduces a minor increase in memory occupied by gradients. This enables tuning much more parameters for PEFT methods.

\subsection{Stabilize Training with LOMO}
\label{sec:stablize_training}
\subsubsection{Alternatives to Gradient Normalization and Clipping}

Gradient normalization and clipping are essential tools to deal with the gradient explosion and vanishing problem~\cite{DBLP:conf/icml/ChenBLR18}, but their computation process requires using the gradient tensors of all parameters. We propose two alternatives here:

\begin{itemize}
    \item Clipping gradient tensors by its values rather than the norm. 
    \item Compute the gradient norm in an additional backward pass. 
\end{itemize}

Clipping gradient tensors by their values is a simple but effective solution for gradient explosion before gradient norm approaches. The main concern of clipping by values is that truncating some gradient elements could change the direction of the gradient tensor. For example, a two-dim vector $[1.3, 0.8]$ and its clipped version $[1.0, 0.8]$ (clipped to $1.0$) indicate different directions. Our experience is that the clipping by values performs worse when the learning rate is high because truncations happened more often in that case. However, clipping by values performs well for medium and small learning rates. Note that the scale of the learning rate largely depends on the task and data, but in general, we suggest using clipping by values for a learning rate less than $1e-3$. 

Our approach cannot directly compute the gradient norm because we update parameters along with the backpropagation, so we do not know the norm of rest parameters when updating a certain parameter. However, we can introduce an additional pass to compute and accumulate each parameter's gradient norm, resulting in two backward passes, one for computing the gradient norm and one for updating parameters. The memory usage leaves unchanged but sacrifices the speed. 

\paragraph{A controversial solution}
Our current training framework computes the gradient norm based on all parameters and requires two backward passes. 
One solution to save the additional backward pass is to approximate the norm of gradient tensors with a group of parameters, for example, the adjacent layers. This method is indeed biased, because it results in different update step sizes for different parameters. When updating, the parameters are multiplied by a scale factor according to the gradient norms. 
Since the gradient norms differ among parameter groups, such an approximation leads to a difference in scale factors. 
This grouped gradient clipping method can be considered as applying a dynamic learning rate to different groups of parameters based on their gradient norms. \citet{DBLP:journals/tcyb/SunCZZ20} suggests that it is not always appropriate to use the same learning rate for all parameters in SGD, thus we believe our approach also holds the potential to further benefit SGD.
We leave the exploration as a compelling future direction.

\subsubsection{Mitigating Precision Degradation}

Mixed-precision training is commonly employed to speed up the training process. To mitigate the degradation in precision, we utilize dynamic loss scaling and transition certain computations to full precision. The approach of loss scaling is crucial in preventing underflows during FP16 training, magnifying the loss with a specific factor prior to the backward pass and diminishing the gradients by the same factor.

In this context, we integrate a dynamic loss scaler with LOMO, which dynamically adjusts the scaling factor throughout the training procedure. If no overflow occurs during a specified number of backward passes, the scale factor is doubled. Otherwise, this step is dropped and the scale factor is halved.
This process echoes the scenario encountered during gradient normalization. It is unknown whether there will be an overflow until the backward has completed. Consequently, we perform two backward passes: the first pass to identify any overflow, and the second pass to update the parameters if no overflow is detected. These two backward passes for dynamic loss scaling can be executed simultaneously with gradient normalization.
To effectively update parameter and handle gradient for operations like normalization and scaling, the gradient and its associated parameter are converted to full precision within these computations.

\begin{figure*}
    \centering
    \includegraphics[width=1.0\textwidth]{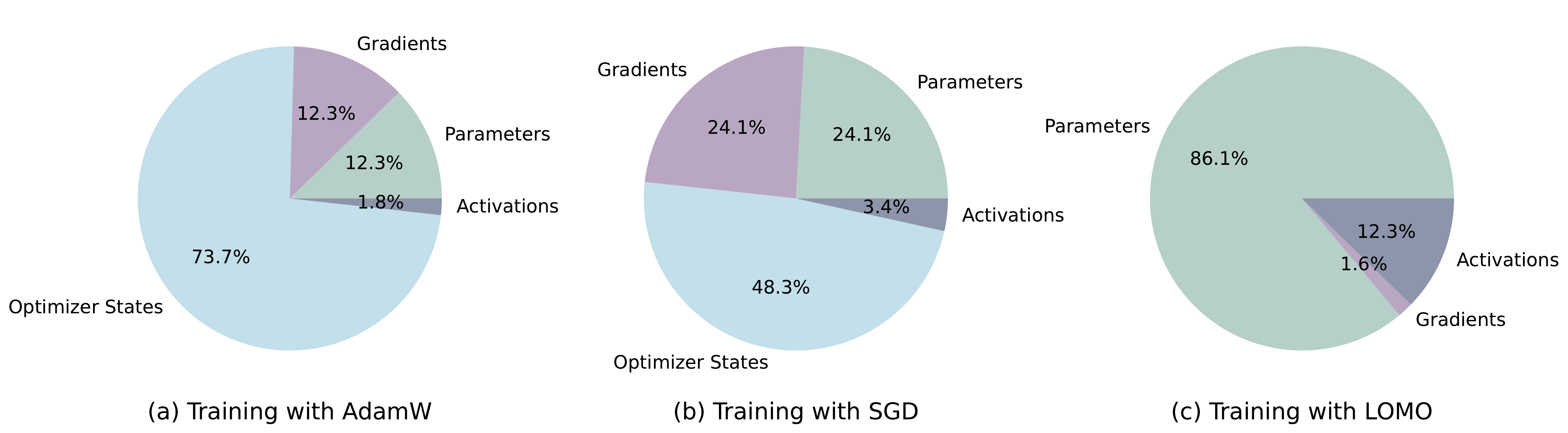}
    \caption{The memory usage ratio of each part when using different optimizers to train LLaMA-7B. The sequence length and batch size are set to 512 and 8, respectively.}
    \label{fig:mem_profile}
\end{figure*}
\begin{table*}[]
\centering
\begin{tabular}{@{}ccccccc@{}}
\toprule
                       & \textbf{AC} & \textbf{Params}                 & \textbf{Gradients}              & \textbf{Optim States}       & \textbf{Activations} & \textbf{Total Memory}  \\ \midrule
\multirow{2}{*}{AdamW} & \ding{55}  & \multirow{2}{*}{12.55} & \multirow{2}{*}{12.55} & \multirow{2}{*}{75.31} & 45.61       & 147.02 \\
                       & \ding{51}  &                        &                        &                        & \ \ 1.79        & 102.20 \\
\midrule
\multirow{2}{*}{SGD}   & \ding{55}  & \multirow{2}{*}{12.55} & \multirow{2}{*}{12.55} & \multirow{2}{*}{25.10} & 45.61       & \ \ 96.81 \\
                       & \ding{51}  &                        &                        &                        & \ \ 1.79        & \ \ 51.99  \\
\midrule
\multirow{2}{*}{LOMO} & \ding{55}  & \multirow{2}{*}{12.55} & \multirow{2}{*}{\ \ 0.24}  & \multirow{2}{*}{\ \ 0.00}     & 45.61       & \ \ 59.40  \\
                       & \ding{51}  &                        &                        &                        & \ \ 1.79        & \ \ 14.58  \\ \bottomrule
\end{tabular}
\caption{Memory usage (GB) when training LLaMA-7B under different settings. \textbf{AC} refers to Activation Checkpointing. The sequence length and batch size are set to 512 and 8, respectively.}
\label{tab:mem_profile}
\end{table*}
\section{Experiment}
In this section, we evaluate our proposed method from three aspects, namely memory profile, throughput and downstream performance.
If not further explained, all our experiments are conducted with LLaMA models~\citep{DBLP:journals/corr/abs-2302-13971}, ranging from 7B to 65B.

\subsection{Memory Profile}

We first profile the memory usage of model states and activations during the training under different settings. As demonstrated in Table~\ref{tab:mem_profile}, the usage of the LOMO optimizer leads to a substantial reduction in memory footprint from 102.20GB to 14.58GB, when compared to the AdamW optimizer~\citep{DBLP:conf/iclr/LoshchilovH19}, and from 51.99GB to 14.58GB, when compared to SGD, in the context of training the LLaMA-7B model. This significant decrease in memory usage can be attributed primarily to the reduced memory requirements of the gradient and optimizer states. As a result, memory is mostly occupied by parameters in the training process, commensurate with memory usage during inference.

\paragraph{Optimizer States} Figure~\ref{fig:mem_profile} illustrates that employing the AdamW optimizer for LLaMA-7B training, a widely adopted configuration, yields a substantial proportion of memory (73.7\%) being allocated to optimizer states. This outcome is a consequence of the mixed-precision training approach, where full-precision copies of weights, momentum, and variance are maintained within the optimizer states for weight updates. Replacing the AdamW optimizer with the SGD optimizer can effectively reduce the percentage of  optimizer states in memory, and therefore alleviate the GPU
memory usage (from 102.20GB to 51.99GB). This reduction is due to the fact that the SGD optimizer does not require the storage of full-precision momentums and variances. For LOMO, parameter update and backward are fused into one step, further eliminating the need for optimizer state memory.
\begin{table*}[]
\centering
\begin{tabular}{@{}ccccc@{}}
\toprule
\textbf{Params} & \textbf{Optimizer} & \textbf{Hardware} & \textbf{Memory (GB)} & \textbf{Throughput (TGS)} \\ \midrule
7B     & AdamW     & 8 $\times$ RTX 3090          &  15.76 &  \ \ 67.37    \\
7B     & SGD       & 8  $\times$ RTX 3090         &  \ \ 9.49 &   \ \ 69.66   \\
7B     & LOMO & 1 $\times$ RTX 3090          & 13.61  &  769.92    \\
\midrule
13B    &       SGD & 8 $\times$ RTX 3090          &  15.74 &   \ \ 32.51    \\
13B    & LOMO & 2  $\times$ RTX 3090         & 15.92  &  \ \ 66.19   \\
\midrule
30B    & LOMO & 4  $\times$ RTX 3090         & 19.78  &  \ \ 11.61     \\
\midrule
65B    & LOMO & 8  $\times$ RTX 3090         & 19.18  &  \ \ \ \ 4.93      \\ \bottomrule
\end{tabular}
\caption{Throughput tested on a server with 8 RTX 3090 GPUs. The sequence length and batch size are set to 1024 and 1,
respectively. \textbf{Memory} represents the peak memory allocated per GPU during training. \textbf{Throughput} represents the number of tokens processed by each GPU per second (TGS).}
\label{tab:throughput}
\end{table*}
\paragraph{Gradients} 
During the training process using LOMO, parameters are immediately updated upon receiving gradients, following which the gradients are discarded from memory. 
As a result, the upper bound of gradient memory consumption is determined by the gradient associated with the parameter matrix of greatest magnitude.
This approach considerably reduces memory usage by almost the size of parameters.

\paragraph{Activations} The training of a 7B model with 512$\times$8 tokens in one batch demands a substantial amount of memory for activations. LOMO is compatible with activation memory reduction techniques such as activation checkpointing. By integrating activation checkpointing with LOMO, the memory footprint due to activation can be reduced from 45.61GB to 1.79GB.

\subsection{Throughput}

We evaluate the throughput performance of LOMO compared to AdamW and SGD. The experiments are conduct on a server equipped with 8 RTX 3090 GPUs, interconnected via a PCIe motherboard. The sequence length and batch size are set to 1024 and 1, respectively. Throughput is measured in terms of the number of tokens processed per GPU per second (TGS), and parameter partitioning was achieved using ZeRO-3~\citep{Zero}.

For the 7B model, LOMO demonstrates remarkable throughput, surpassing AdamW and SGD by about 11 times. This significant improvement can be attributed to LOMO's ability to train the 7B model on a single GPU, thereby reducing inter-GPU communication overhead. The slightly higher throughput of SGD compared to AdamW can be attributed to SGD's exclusion of momentum and variance calculations.

As for the 13B model, it could not be trained with AdamW on the available 8 RTX 3090 GPUs due to memory limitations. In this scenario where model parallelism is necessary for LOMO, LOMO still outperforms SGD in terms of throughput. This advantage is attributed to LOMO's memory-efficient properties and the requirement of only two GPUs to train the model with the same settings, resulting in reduced communication costs and greater throughput.
Furthermore, when training the 30B model, SGD encounters out-of-memory (OOM) issues with the 8 RTX 3090 GPUs, while LOMO performs well with only 4 GPUs.

Finally, we successfully train the 65B model using 8 RTX 3090 GPUs, achieving a throughput of 4.93 TGS. Utilizing such a server configuration and LOMO, the training process on 1000 samples, each containing 512 tokens, requires approximately 3.6 hours.

\subsection{Downstream Performance}

To assess the effectiveness of LOMO in fine-tuning large language models, we conduct an extensive set of experiments. We compare LOMO against two other methods, Zero-shot, which does not require fine-tuning, and LoRA, which is currently one of the most popular parameter-efficient fine-tuning techniques. As descirbed in~\citep{DBLP:conf/iclr/HuSWALWWC22}, LoRA reparameterizes the dense layers and only updates low rank matrices while introducing no latency during inference.

We use the SuperGLUE dataset collection to evaluate model performance, specifically focusing on RTE~\citep{DBLP:conf/mlcw/DaganGM05}, BoolQ~\citep{DBLP:conf/naacl/ClarkLCK0T19}, WSC~\citep{DBLP:conf/kr/LevesqueDM12}, WIC~\citep{DBLP:conf/naacl/PilehvarC19}, MultiRC~\citep{DBLP:conf/naacl/KhashabiCRUR18}, and COPA~\citep{DBLP:conf/aaaiss/RoemmeleBG11}. Given the high computational cost associated with running large language models, we follow MeZO~\citep{DBLP:journals/corr/abs-2305-17333} to randomly sample 1000 training data from training set and 1000 test data from validation set, and report the best results obtained using the same random seed. The prompts used in our experiments are the same as MeZO, and the hyperparameters are detailed in Appendix-\ref{sec:hyparams}.

During inference, we insert different labels or candidates into the prompt and calculate the average log-likelihood for each label. The label with the highest score is selected as the model's answer. To evaluate the performance, we use Accuracy as the evaluation metric.

\subsubsection{Main results}
\begin{table*}[t]
\centering
    \begin{tabular}{lrcccccccc}
    \toprule
     \textbf{Method}  & \textbf{Params} & \textbf{RTE} & \textbf{BoolQ} & \textbf{WSC} & \textbf{WIC}	& \textbf{MultiRC} & \textbf{COPA} & \textbf{Avg.} \\
    \midrule
    Zero-shot & 7B & 57.0 & 66.5 & 36.5 & 49.7 & 42.3 & 85.0 & 56.2\\
    LoRA & 7B & 85.9 & 85.2 & 64.4 & 65.5 & \textbf{84.8} & 87.0 & 78.8\\
    LOMO & 7B & \textbf{86.6} & \textbf{87.5} & \textbf{66.4} & \textbf{71.2} & 84.0 & \textbf{89.0} & \textbf{80.8}\\
    \midrule
    Zero-shot & 13B & 60.6 & 65.0 & 36.5 & 49.5 & 43.4 & 88.0 & 57.2\\
    LoRA & 13B & 89.9 & 87.1 & 63.5 & 69.9 & \textbf{86.1} & 92.0 & 81.4\\
    LOMO & 13B & 89.9 & \textbf{87.3} & \textbf{75.0} & \textbf{74.3} & 85.7 & \textbf{93.0} & \textbf{84.2}\\
    \midrule
    Zero-shot & 30B & 53.4 & 74.6 & 36.5 & 50.0 & 46.9 & 89.0 & 58.4\\
    LoRA & 30B & 91.0 & \textbf{89.7} & 83.7 & 74.0 & 87.0 & 93.0 & 86.4\\
    LOMO & 30B & \textbf{92.8} & 89.3 & \textbf{85.6} & \textbf{74.1} & \textbf{87.9} & 93.0 & \textbf{87.1}\\
    \midrule
    Zero-shot & 65B & 59.6 & 73.6 & 44.2 & 51.3 & 48.3 & 91.0 & 61.3\\
    LoRA & 65B & 93.1 & \textbf{90.9} & 88.5 & 74.5 & \textbf{90.0} & 97.0 & 89.0\\
    LOMO & 65B & \textbf{93.9} & 90.7 & \textbf{92.3} & \textbf{75.4} & 89.9 & 97.0 & \textbf{89.9}\\
    
    \bottomrule
    \end{tabular}
    \caption{
        Main results on SuperGLUE using LLaMA at all sizes (with 1,000 training examples).
    }
    \label{tab:main-results}
\end{table*}

The downstream performances of LOMO compared with Zero-shot and LoRA are presented in Table~\ref{tab:main-results}. Based on the results, we reach the following observations.

\textbf{LOMO performs significantly better than Zero-shot.} Across all six datasets and model sizes, LOMO consistently achieves superior results over Zero-shot, with average gains of more than 20 points using LLaMA-13B. While previous research has showcased the impressive capabilities of large language models in zero-shot settings, fine-tuning still yields remarkable performance enhancements for specific downstream tasks. The experimental results confirm the effectiveness of LOMO in optimizing large language models of different sizes.

\textbf{LOMO generally outperforms LoRA in most experiments.} We show that LOMO delivers strong performance  compared to LoRA, for instance, resulting in average gains of 2.8 points using LLaMA-13B. This suggests that the model performance benefits more from full-parameter fine-tuning than parameter-efficient fine-tuning, as the former adjusts more parameters. LOMO strikes a good balance between performance and efficiency, making it a competitive choice for fine-tuning.

In some cases, LOMO performs worse than LoRA. One possible reason is the relatively small training set we use, which may not be sufficient for full-parameter fine-tuning of large models. Additionally, LoRA and LOMO employ different model architectures. To be specific, LoRA offers a shortcut for model tuning, which can be advantageous in certain scenarios. Actually, these two methods are not conflicting or mutually exclusive. In the next subsection, we validate that combing LoRA with LOMO does not harm model performance and, in most cases, leads to performance gains.

\textbf{LOMO efficiently scales up to 65 billion parameter models.} Despite conducting all experiments on a single machine equipped with 8 $\times$ RTX 3090, LOMO consistently exhibits strong performance even on a 65-parameter scale. This further supports the effectiveness of LOMO in optimizing LLMs under resource-constrained scenarios.

\subsubsection{LoRA with LOMO}
\begin{figure*}[!t]
    \centering
    \includegraphics[width=0.8\textwidth]{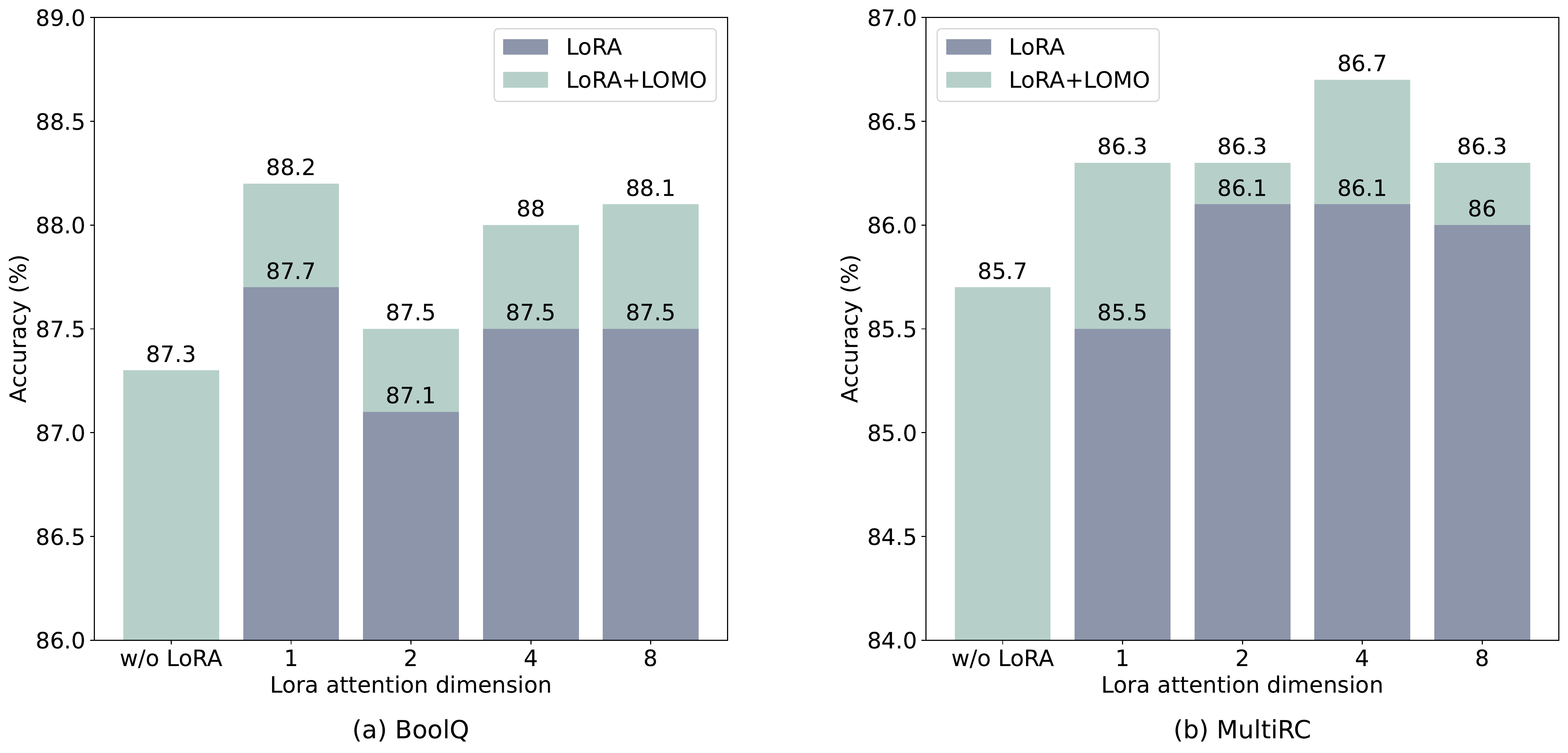}
    \caption{Results using LLaMA-13B on the BoolQ and MultiRC datasets (with 1,000 training examples). ``LoRA+LOMO" means injecting LoRA modules while fine-tuning the pre-trained model weights using LOMO.}
    \label{fig:lora_mesgd}
\end{figure*}
LOMO and LoRA are fundamentally independent of each other. In order to verify this claim, we perform experiments using LLaMA-13B on the BoolQ and MultiRC datasets. Results are shown in Figure~\ref{fig:lora_mesgd}. We find that 
LOMO consistently enhances the performance of LoRA regardless of the higher results LoRA achieved. This suggests that different fine-tuning methods employed by LOMO and LoRA are complementary. Specifically, LOMO focuses on fine-tuning the pre-trained models weights, while LoRA tunes additional modules. As a result, LOMO dose not compromise the performance of LoRA; rather, it facilitates better model tuning for downstream tasks.

\section{Conclusion}
In this paper, we introduce LOw-Memory Optimization (LOMO), a new optimizer designed to facilitate full parameter fine-tuning for large language models with limited resources. We have demonstrated the feasibility of fine-tuning a 65B model on a server equipped with consumer GPUs such as RTX 3090. By analyzing the memory usage of LOMO, conducting throughput tests, and performing experiments on SuperGLUE, we have showcased its effectiveness and potential impact.

Looking ahead, our future work aims to further lower the resource threshold required for training large language models, thus enabling wider access and adoption of these models. 
The majority of memory are currently occupied by parameters when training with LOMO. Thus, one promising direction is the exploration of parameter quantization techniques, which could significantly reduce memory usage. Additionally, we intend to investigate more applicable scenarios for LOMO and delve into theoretical analyses for optimizing large language models, which hold substantial value for advancing the field.

\section*{Limitations}
In response to the challenges associated with gradient normalization and clipping, we have developed alternative optimization methods. Although gradient normalization for LOMO does not increase memory usage, our current implementation necessitates an additional backward pass, which can slow down the training speed in scenarios where gradient normalization is essential. 

Due to time and resource constraints, our experiments were limited to a subset of the SuperGLUE benchmark, and we did not evaluate LOMO's throughput on advanced GPUs such as A100.

\section*{Ethics statement}
This paper employs open-source models LLaMA, in compliance with their respective licenses. The datasets utilized, including RTE, BoolQ, WSC, WIC, MultiRC and COPA, permit public and freeusage.

\section*{Acknowledgments}
This work was supported by the National Key Research and Development Program of China (No.2022ZD0160102). The computations in this research were performed using the CFFF platform of Fudan University.

\bibliography{custom}

\appendix

\section{Hyperparameters}
\label{sec:hyparams}

The hyperparameters we use in the experiments are listed in Table~\ref{tab:hyperparams}. Due to limited computational resources, we report the highest results of experiments conducted with the same random seed.

\begin{table*}[h]
\centering
    \begin{tabular}{lcc}
    \toprule
    \textbf{Experiments} & \textbf{Hyperparameters} & \textbf{Values} \\
    \midrule
    & LR Schedule & Linear\\
    & Max Grad Norm & 1.0\\
    & Batch size & 16\\
    & \# Epochs & 10\\
    \midrule
    \multirow{2}{*}{LOMO} & Learning Rate & \{5e-2, 3e-2\}\\
    & Warmup Ratio & \{0.05, 0.1, 0.2\}\\
    \midrule
    \multirow{5}{*}{LoRA} & Optimizer & AdamW\\
    & Learning Rate & 5e-4\\
    & Warmup Ratio & 0.05\\
    & LoRA Config. & $r_q = r_v = 2$\\
    & LoRA $\alpha$ & 16\\
    \midrule
    \multirow{7}{*}{LoRA+LOMO} & LoRA Optimizer & AdamW\\
    & LoRA Learning Rate & 5e-4\\
    & LoRA Warmup Ratio & 0.05\\
    & LoRA Config. & $r_q = r_v = \{1, 2, 4, 8\}$\\
    & LoRA $\alpha$ & 16\\
    \cmidrule{2-3}
    & LOMO Learning Rate & \{5e-3, 1e-3, 5e-4\}\\
    & LOMO Warmup Ratio & {0.05, 0.1}\\
    \bottomrule
    \end{tabular}
    \caption{The hyperparameters used in our experiments.}
    \label{tab:hyperparams}
\end{table*}

\section{Training Dynamics}
\begin{figure}
    \centering
    \includegraphics[width=\linewidth]{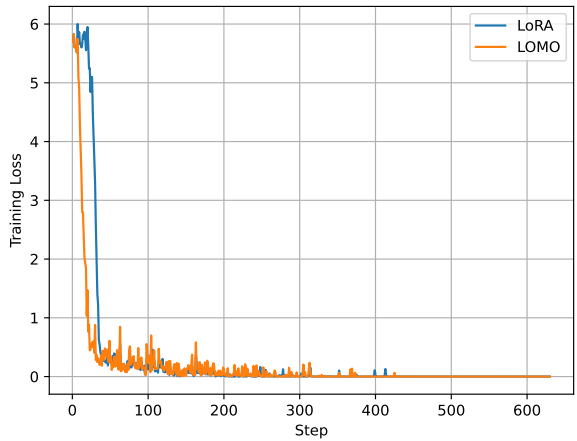}
    \caption{Training loss curve for LLaMA-7B on BoolQ.}
    \label{fig:loss_curve}
\end{figure}

\begin{figure}
    \centering
    \includegraphics[width=\linewidth]{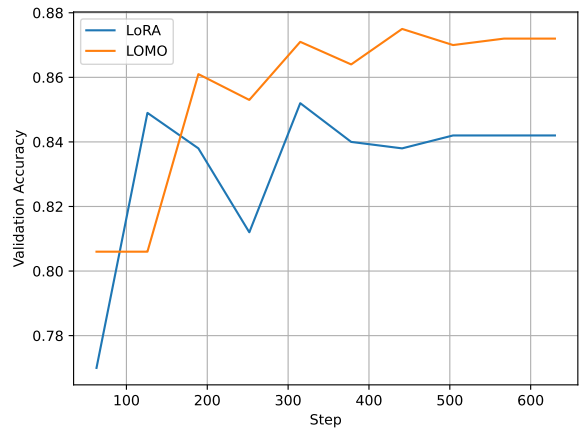}
    \caption{Validation accuracy of LLaMA-7B on BoolQ.}
    \label{fig:valid_acc}
\end{figure}

To analyze the training dynamics of LOMO, we present the training loss curve and validation accuracy for LLaMA-7B trained on BoolQ~\citep{DBLP:conf/naacl/ClarkLCK0T19} using LOMO and LoRA in Figure~\ref{fig:loss_curve} and Figure~\ref{fig:valid_acc}, respectively. 
During training process with LOMO, the loss converges rapidly in the initial phase and then tends to stabilize and gradually decline. The accuracy on the development set generally shows an upward trend as the number of training steps increases.

\end{document}